\documentclass{article}
\global\long\def\hiddenScalar{\mappingFunction}
\global\long\def\numLayers{\ell}
\global\long\def\conditionalCovariance{\boldsymbol{\Sigma}}
\global\long\def\shortlong#1#2{\ifdefined\shortVersion #1\else #2\fi}
\newcommand{\given}{|}

\shortlong{\usepackage{aistats2015}}{}

\usepackage{amsmath, amssymb}
\usepackage{natbib}

\usepackage{subcaption}
\usepackage{graphicx}
\newcommand{\todiagrams}{}
\usepackage{tikz}
\usetikzlibrary{bayesnet}
\usepackage{pgfplots}
\usepgfplotslibrary{groupplots}
\pgfplotsset{compat=newest}
\usepgfplotslibrary{external} 
\newlength\figureheight
\newlength\figurewidth
\usepackage{algorithm}
\usepackage{algorithmic}

\usepackage{todonotes}

\usepackage{color}
\usepackage{verbatim}

\definecolor{brown}{rgb}{0.9,0.59,0.078}
\definecolor{ironsulf}{rgb}{0,0.7,.5}
\definecolor{lightpurple}{rgb}{0.156,0,0.245}

\ifdefined\blackBackground
\definecolor{colorOne}{rgb}{0, 1, 1}
\definecolor{colorTwo}{rgb}{1, 0, 1}
\definecolor{colorThree}{rgb}{1, 1, 0}
\definecolor{colorTwoThree}{rgb}{1, 0, 0}
\definecolor{colorOneThree}{rgb}{0, 1, 0}
\definecolor{colorOneTwo}{rgb}{0, 0, 1}
\else
\definecolor{colorOne}{rgb}{1, 0, 0}
\definecolor{colorTwo}{rgb}{0, 1, 0}
\definecolor{colorThree}{rgb}{0, 0, 1}
\definecolor{colorTwoThree}{rgb}{0, 1, 1}
\definecolor{colorOneThree}{rgb}{1, 0, 1}
\definecolor{colorOneTwo}{rgb}{1, 1, 0}
\fi

\ifdefined\blackBackground

\else

\fi

\global\long\def\det#1{\left|#1\right|}

\global\long\def\refeq#1{(\ref{#1})}

\global\long\def\eye{\mathbf{I}}

\global\long\def\trace#1{\text{tr}\left(#1\right)}

\global\long\def\cut#1{}

\global\long\def\reffig#1{figure~\ref{#1}}
\global\long\def\Reffig#1{Figure~\ref{#1}}

\global\long\def\detail#1{}

\global\long\def\conditionalCovariance{\boldsymbol{\Sigma}}

\global\long\def\numLayers{\ell}
\global\long\def\numData{n}

\global\long\def\lengthScale{\ell}

\global\long\def\dataScalar{y}
\global\long\def\dataVector{\mathbf{\dataScalar}}

\global\long\def\bigO{\mathcal{O}}

\global\long\def\hiddenScalar{h}

\global\long\def\hiddenVector{\mathbf{\hiddenScalar}}

\global\long\def\noiseVector{\boldsymbol{\epsilon}}

\global\long\def\kernelScalar{k}

\global\long\def\kernelMatrix{\mathbf{\MakeUppercase{\kernelScalar}}}

\global\long\def\lengthScale{\ell}

\global\long\def\zerosVector{{\bf 0}}

\global\long\def\numData{n}

\global\long\def\inputScalar{x}
\global\long\def\inputMatrix{{\bf \MakeUppercase{\inputScalar}}}
\global\long\def\inputVector{{\bf \inputScalar}}

\global\long\def\meanScalar{\mu}

\global\long\def\mappingFunction{f}
\global\long\def\mappingFunctionVector{\mathbf{\mappingFunction}}

\global\long\def\mappingFunctionTwo{g}
\global\long\def\mappingFunctionTwoVector{\mathbf{\mappingFunctionTwo}}

\global\long\def\inducingScalar{u}
\global\long\def\inducingVector{\mathbf{\inducingScalar}}

\global\long\def\inducingInputScalar{z}
\global\long\def\inducingInputVector{\mathbf{\inducingInputScalar}}
\global\long\def\inducingInputMatrix{\mathbf{\MakeUppercase{\inducingInputScalar}}}

\global\long\def\basisFunction{\phi}
\global\long\def\basisVector{\boldsymbol{\basisFunction}}

\global\long\def\eye{\mathbf{I}}

\global\long\def\zerosVector{\mathbf{0}}
\global\long\def\half{\frac{1}{2}}

\global\long\def\gaussianSamp#1#2{\mathcal{N}\left(#1,#2\right)}
\global\long\def\gaussianDist#1#2#3{\mathcal{N}\left(#1|#2,#3\right)}

\global\long\def\expSamp#1{\left<#1\right>}
\global\long\def\expDist#1#2{\left<#1\right>_{#2}}

\global\long\def\KL#1#2{\text{KL}\left( #1\,\|\,#2 \right)}

\global\long\def\kff{\kernelScalar_{\mappingFunction \mappingFunction}}

\global\long\def\Kuu{\kernelMatrix_{\inducingVector \inducingVector}}

\global\long\def\Kfu{\kernelMatrix_{\mappingFunctionVector \inducingVector}}
\global\long\def\Kuf{\kernelMatrix_{\inducingVector \mappingFunctionVector}}

\global\long\def\det#1{\left|#1\right|}

\global\long\def\ltwoNorm#1{\left\Vert #1 \right\Vert_2}

\begin{document}

\shortlong
{
\twocolumn[
\aistatstitle{Nested Variational Compression in  Deep Gaussian Processes}
\aistatsauthor{ James Hensman \And Neil D. Lawrence}
\aistatsaddress{ Department of Computer Science, University of Sheffield, UK}
}
{
\title{Nested Variational Compression in Deep Gaussian Processes}
\author{James Hensman and Neil D. Lawrence}
\maketitle
}

\begin{abstract} 
Deep Gaussian processes provide a flexible approach to probabilistic
modeling of data using either supervised or unsupervised learning. For
tractable inference approximations to the marginal likelihood of the
model must be made. The original approach to approximate inference in
these models used variational compression to allow for approximate
variational marginalization of the hidden variables leading to a lower
bound on the marginal likelihood of the model
\citep{Damianou:deepgp13}. In this paper we extend this idea with a
\emph{nested} variational compression. The resulting lower bound on
the likelihood can be easily parallelised or adapted for stochastic
variational inference.

\end{abstract}

\section{Introduction}
\label{sec:intro}

Gaussian process (GP) models provide flexible non-parameteric probabilistic
approaches to function estimation in an analytically tractable
manner. However, their tractability comes at a price: they can only
represent a restricted class of functions. \shortlong{GPs}{Gaussian processes} came to
the attention of the machine learning community through the PhD thesis
of Radford Neal, later published in book form \citep{Neal:book96}. At
the time there was still a great deal of interest in the result that a
neural network with a single layer and an infinite number of hidden
units was a universal approximator \citep{Hornik:univ89}, but Neal was
able to show that in such a limit the model became a \shortlong{GP}{Gaussian process}
with a particular covariance function (the form of the covariance
function was later derived by \citet{Williams:computation98}). Both
\citet{Neal:book96} and \citet{MacKay:gpintroduction98} pointed out
some of the limitations of priors that ensure joint Gaussianity across
observations and this has inspired work in moving beyond Gaussian
processes \citep{Wilson:gpatt13}.

\shortlong{\paragraph{Process Composition}}{\subsection{Process Composition}}
Deep Gaussian processes
\citep{Lawrence:hgplvm07,Damianou:vgpds11,Lazaro:warped12,Damianou:deepgp13}
are an attempt to address this limitation.

In a deep \shortlong{GP}{Gaussian process}, rather than assuming that a data
observation, $\dataVector$, is a draw from a \shortlong{GP}{Gaussian process} prior,
\shortlong{$}{\[}
\dataVector = \mappingFunction(\inputVector) + \noiseVector,
\shortlong{$}{\]}
where $\mappingFunction(\inputVector)$ is a vector-valued function
drawn from a \shortlong{GP}{Gaussian process} and $\noiseVector$ is a noise
corruption, we make use of a functional composition,
\shortlong{$}{\[}
\dataVector=\mappingFunctionVector_\ell(\mappingFunctionVector_{\ell-1}(\dots \mappingFunctionVector_1(\inputVector))) + \noiseVector,
\shortlong{$}{\]}
where we assume each function in the composition,
$\mappingFunctionVector_i(\cdot)$, is itself a draw from a Gaussian
process. In other words we develop our full probabilistic model through
\emph{process composition}.

Process composition has the appealing property of retaining the
theoretical qualities of the underlying stochastic process (such as
Kolmogorov consistency) whilst providing a richer class of process
priors. For example, for deep \shortlong{GPs}{Gaussian processes}
\citet{Duvenaud:pathologies14} have shown that, for particular
assumptions of covariance function parameters, the derivatives of
functions sampled from the process have a marginal distribution that
is heavier tailed than a Gaussian. In contrast, it is known that in a
standard \shortlong{GP}{Gaussian process} the derivatives of functions sampled from
the process are jointly Gaussian with the original function.

\section{Deep Models}

Process composition in \shortlong{GPs}{Gaussian processes} has become known as \emph{deep \shortlong{GPs}{Gaussian processes}} due to the relationship between these models and deep neural network models. A single layer neural network has the following form,
\shortlong{$}{\[}
\mappingFunctionTwoVector(\inputVector) = \mathbf{V}^\top \basisVector(\mathbf{U}\inputVector)
\shortlong{$}{\]}
where $\basisVector(\cdot)$ is a vector valued function of an adjustable basis, which is controlled by a parameter matrix $\mathbf{W}$, and $\mathbf{V}$ is used to provide a linear weighted sum of the basis to give us the resulting vector valued function, in a similar way to generalised linear models. Deep neural networks then take the form of a functional composition for the basis functions,
\shortlong{$}{\[}
\mappingFunctionTwoVector (\inputVector)= \mathbf{V}_\numLayers^\top\basisVector_{\numLayers}(\mathbf{W}_{\numLayers-1}\basisVector_{\numLayers-1}(\dots\mathbf{W}_2\basisVector(\mathbf{U}_1\inputVector)).
\shortlong{$}{\]}

A serious challenge for deep networks, when trained in a feed-forward manner, is overfitting. As the number of layers increase, and the number of basis functions in each layer also goes up a very powerful representation that is highly parameterised is formed. The matrix mapping between each set of basis functions $\mathbf{W}_i$ has size $k_i \times k_{i+1}$, where $k_i$ is the number of basis functions in the $k$th layer. In practice networks containing sometimes thousands of basis functions can be used leading to a parameter explosion.

\shortlong{\paragraph{Weight Matrix Factorization}}{\subsection{Weight Matrix Factorization}}
One approach to dealing with such matrices is to replace them with a lower rank form,
$$
\mathbf{W}_i = \mathbf{U}_i\mathbf{V}_i^\top
$$
where $\mathbf{U}_i\in \Re^{k_{i+1} \times q_i}$ and $\mathbf{V} \in
  \Re^{k_{i} \times q_i}$, where $q_i<k_i$ and $q_i<k_{i+1}$. Whilst
  this idea hasn't yet, to our knowledge, been yet pursued in the deep
  neural network community, \citet{Denil:predicting13} have
  empirically shown that trained neural networks can have low rank
  matrices as evidenced by the ability to predict one set of part of
  the weight matrix given by another. The approach of `dropout'
  \citep{Srivastava:dropout14} is also widely applied to control the
  complexity of the model implying the models are over parameterised.

Substituting the low rank form into the compositional structure for
the deep network we have
\shortlong{$}{\[}
\mappingFunctionTwoVector(\inputVector)=
\mathbf{V}_\numLayers^\top\basisVector_{\numLayers}(\mathbf{U}_{\numLayers-1}\mathbf{V}_{\numLayers-1}^\top\basisVector_{\numLayers-1}(\dots(\mathbf{U}_{2}\mathbf{V}_2\basisVector(\mathbf{U}_1\inputVector)).
\shortlong{$}{\]}
We can now identify the following form inside the functional decomposition,
\shortlong{$}{\[}
\mappingFunction_i(\mathbf{z}) = \mathbf{V}_i^\top
\basisVector_i(\mathbf{U}_{i-1}\mathbf{z})
\shortlong{$}{\]}
and once again obtain a functional composition,
\shortlong{$}{\[}
\mappingFunctionTwoVector(\inputVector)=\mappingFunctionVector_\ell(\mappingFunctionVector_{\ell-1}(\dots
\mappingFunctionVector_1(\inputVector))).
\shortlong{$}{\]}
The standard deep network is recovered when we set $q_i =
\text{min}(k_i, k_{i+1})$ and the deep \shortlong{GP}{Gaussian process} is recovered
by keeping $q_i$ finite and allowing $k_i \rightarrow \infty$ for all
layers. Of course, the mappings in the \shortlong{GP}{Gaussian process} are treated
probabilistically and integrated out, rather than optimised, so
despite the increase in layer size the number of parameters in the
resulting model is many fewer than those in a standard deep neural
network.

Deep \shortlong{GPs}{Gaussian processes} can also be used for unsupervised learning by
replacing the input nodes with a white noise process. The resulting
deep GPs can be seen as fully Bayesian generalizations of unsupervised
deep neural networks with Gaussian nodes as proposed by
\citet{Kingma:auto13} and \citet{Rezende:stochastic14}.

Unfortunately, exact inference in deep \shortlong{GP}{Gaussian process} models is
intractable. \citet{Damianou:deepgp13} applied variational compression
\citep{Titsias:variational09} to perform approximate inference. In this
paper we revisit that bound and apply a nested variational compression
to obtain a new bound on the deep GP. The new bound factorizes across
data points allowing parallel computation \citep{Gal:distributed14} or
optimization via stochastic gradient descent
\citep{Hensman:bigdata13}. 

\shortlong{}{ In the rest of the paper, we first review variational
  compression in the context of Gaussian processes. We then introduce
  deep Gaussian processes and show how the compression can be applied
  in a nested way to perform inference through an arbitrary number of
  model layers. We end with experiments on some standard data sets.  }

\section{Variational Compression in \shortlong{GPs}{Gaussian Processes}}
\label{sec:gp}

Let's assume that we are given a series of input-output pairs $\{\dataScalar_i,
\inputVector_i\}_{i=1}^\numData$, which we stack into target vector and design
matrix $\dataVector, \inputMatrix$. The data are modelled as noisy observations
of a function $f$, so that
\begin{equation}
\dataScalar_i = \mappingFunction(\inputVector_i) + \epsilon_i,
\end{equation}
with $\epsilon_i \sim \gaussianSamp{0}{\sigma^2}$. In a \shortlong{GP}{Gaussian process} model we assume that the function $\mappingFunction(\inputVector)$ is drawn from a \shortlong{GP}{Gaussian process},
$
\mappingFunction(\inputVector) \sim \mathcal{GP}(\meanScalar_\mappingFunction(\inputVector), \kff(\inputVector, \inputVector^\prime)),
$ 
by which we mean that any finite set of values of the function will be jointly Gaussian distributed with a mean given by computing $\meanScalar(\cdot)$ at the relevant points and a covariance given by computing $\kff(\cdot,\cdot)$ at the relevant points. 

The consistency property of the \shortlong{GP}{Gaussian process} enables us to
consider the values of the function on where the data are present,
effectively marginalising the remaining values. If we assume that the
mean function is zero, then we can write
\begin{equation}
  p(\mappingFunctionVector\given\inputMatrix) = \gaussianDist{\mappingFunctionVector}{\bf 0}{\kernelMatrix_{\mappingFunctionVector\mappingFunctionVector}}.
    \label{eq:gp_consistent}
\end{equation}
The beauty of \shortlong{GPs}{Gaussian processes} lies in their tractability. Using
properties of multivariate Gaussians it is straightforward to compute
the marginal likelihood (in $\bigO(\numData^3)$ complexity), as well
as the posterior of the latent function values
$p(\mappingFunctionVector\given\dataVector, \inputMatrix)$ \citep[see
  for example][]{Rasmussen:book06}. Note that the covariance function,
$\kff(\cdot, \cdot)$, or kernel, is also dependent on
parameters that control properties (such as lengthscale, smoothness
etc). We omit this dependence in our notation, and we assume that such
parameters might be determined as an outer loop on our algorithm such
as maximum (approximate) likelihood or an appropriate approximate
Bayesian procedure.

\shortlong{\paragraph{Inducing Points}}{\subsection{Inducing Points Representations}}
Gaussian processes are flexible non-parametric models for
functions. However, whilst inference is analytically tractable, a
challenge for these models lies in their computational complexity and
storage which are worst case $\bigO(\numData^3)$ and
$\bigO(\numData^2)$ respectively.

To deal with this, in the machine learning community, there has been a lot of focus on augmenting \shortlong{GP}{Gaussian process} models by introducing an extra set of variables, $\inducingVector$, and their corresponding inputs, $\inducingInputMatrix$ \citep[see e.g.][]{Csato:sparse02,Snelson:pseudo05,Quinonero:unifying05}. Augmentation occurs by assuming that there is an additional set of variables $\inducingVector$ which are jointly Gaussian with our original function $\mappingFunctionVector$ allowing us to write 
\begin{equation}
p(\mappingFunctionVector, \inducingVector\given\inducingInputMatrix, \inputMatrix) = \gaussianDist{\left[\begin{array}{c} \mappingFunctionVector\\ \inducingVector\end{array}\right]}
{{\mathbf 0}}
{\left[\begin{array}{cc}
\kernelMatrix_{\mappingFunctionVector\mappingFunctionVector} & \kernelMatrix_{\mappingFunctionVector\inducingVector}\\
\kernelMatrix_{\inducingVector\mappingFunctionVector} & \kernelMatrix_{\inducingVector\inducingVector}
\end{array}\right]}
\end{equation}
The multivariate joint Gaussian density has the convenient property
that it is trivial to decompose it into associated conditional and
marginal distributions for each variable. We can decompose the joint
distribution as follows
\begin{align}
  p(\mappingFunctionVector, \inducingVector) &=
  p(\mappingFunctionVector\given
  \inducingVector)p(\inducingVector)\\ &=
  \gaussianDist{\mappingFunctionVector}{\kernelMatrix_{\mappingFunctionVector\inducingVector}\kernelMatrix_{
      \inducingVector\inducingVector
    }^{-1}\inducingVector}{\conditionalCovariance}\gaussianDist{\inducingVector}{\zerosVector}{\kernelMatrix_{
      \inducingVector\inducingVector }}
  \label{eq:augmented_expanded}
\end{align}
where
$
\conditionalCovariance = \kernelMatrix_{\mappingFunctionVector\mappingFunctionVector} - \kernelMatrix_{\mappingFunctionVector \inducingVector }\kernelMatrix_{ \inducingVector\inducingVector }^{-1}\kernelMatrix_{ \inducingVector\mappingFunctionVector}
$ 
is the conditional covariance of $\mappingFunctionVector$ given $\inducingVector$. The model is now augmented with a set of additional latent variables $\inducingVector$. Of course, we can immediately marginalise these variables exactly and return to equation \refeq{eq:gp_consistent}, but through their introduction we will be able to apply a variational approach to inference termed \emph{variational compression}.

Reintroducing independent Gaussian noise we can write the joint density for the data and the augmented latent variables as
\begin{equation}
  p(\dataVector,\mappingFunctionVector,\inducingVector) =
  p(\dataVector\given\mappingFunctionVector)
  p(\mappingFunctionVector\given\inducingVector) p(\inducingVector),
  \label{eq:gp_construction}
\end{equation}
where we have ignored the conditioning on the input locations
$\inputMatrix$.
The technique of
variational compression 
now proceeds by
considering the conditional density
\begin{equation}
  p(\dataVector|\inducingVector) =
  \int p(\dataVector\given\mappingFunctionVector)
  p(\mappingFunctionVector\given\inducingVector)\,\textrm d\mappingFunctionVector
\end{equation}
which we choose to lower bound through assuming an approximation to
the posterior of the form $q(\mappingFunctionVector, \inducingVector)
= p(\mappingFunctionVector|\inducingVector)q(\inducingVector)$
obtaining,
\begin{equation}
  \log p(\dataVector\given\inducingVector, \inducingInputMatrix,
  \inputMatrix) \geq
  \gaussianDist{\dataVector}{\kernelMatrix_{\mappingFunctionVector\inducingVector
    }\kernelMatrix_{\inducingVector\inducingVector}^{-1}\inducingVector}{\sigma^2\eye} - \tfrac{1}{2\sigma^2}\trace{\conditionalCovariance}
\label{eq:gp_conditional}
\end{equation}
This conditional bound now has two parts, a part that looks like a
likelihood conditioned on the inducing variables and a part that acts
as a correction to the lower bound.

\shortlong{\paragraph{Low Rank Approximations}}{\subsection{Low Rank Gaussian Process Approximations}}
Since the conditional bound \refeq{eq:gp_conditional} is conjugate to
the prior $p(\inducingVector)$, we can integrate out the remaining variables
$\inducingVector$. The result is
\begin{equation}
  \log p(\dataVector\given\inputMatrix, \inducingInputMatrix) \geq \log\gaussianDist{\dataVector}{\zerosVector}{\kernelMatrix_{\mappingFunctionVector\inducingVector}\kernelMatrix_{ \inducingVector\inducingVector}^{-1}\kernelMatrix_{\inducingVector\mappingFunctionVector} + \sigma^2\eye} -\tfrac{1}{2\sigma^2}\trace{\conditionalCovariance},
  \label{eq:titsias}
\end{equation}
Note the similarity here between the approximation and a traditional
Bayesian parametric model. In a parametric model we normally have a
likelihood conditioned on some parameters and we integrate over the
parameters using a prior. Our inducing variables appear somewhat
analogous to the parameters in that case. However, before
marginalization the likelihood is independent across the data
points. This observation inspires stochastic variational approaches
\citep{Hoffman:stochastic12} to allow GP inference for very large data
sets \citet{Hensman:bigdata13}.

\shortlong{\paragraph{Stochastic Variational Inference}}{\subsection{Stochastic Variational Inference}}
Direct marginalization of the inducing variables, $\inducingVector$,
is tempting as it is analytically tractable. However it results in an
expression which does not factor in $\numData$, and so does not lend
itself to stochastic optimization. Now if we apply traditional parametric variational Bayes to this portion (treating $\inducingVector$ as the `parameters' of the model) we obtain
\begin{align}
  \log p(\dataVector\given\inputMatrix,\inducingInputMatrix)  \geq &\log \gaussianDist{\dataVector}{\Kfu \Kuu^{-1}\mathbf{m}}{\sigma^2\eye} \nonumber \\
&  - \frac{1}{2\sigma^2} \trace{\mathbf{S} \Kuu^{-1}\Kuf\Kfu\Kuu^{-1}} \nonumber \\
&  - \KL{q(\inducingVector)}{p(\inducingVector)} -\frac{1}{2\sigma^2}\trace{\conditionalCovariance}
  \label{eq:svigp}
\end{align}
where the variational distribution is $q(\inducingVector) = \gaussianDist{\inducingVector}{\mathbf{m}}{\mathbf{S}}$. This method was used by \citet{Hensman:bigdata13} to fit GPs to large datasets through stochastic variational optimization \citep{Hoffman:stochastic12}. 

\shortlong{\paragraph{Total Conditional Variance}}{\subsection{Mutual Information and the Total Conditional Variance}}
\label{par:trace_term}
There are two conditions under which the variational compression will provide a tight lower bound. First, if the data $\dataVector$ are less informative about the latent function variables $\mappingFunctionVector$, and second where the inducing variables $\inducingVector$ are highly informative about $\mappingFunctionVector$. The former depends on the noise variance, $\sigma^2$, and the latter on the conditional entropy of the latent variables given the inducing variables. The conditional entropy for the GP system is given by
\shortlong{$}{\[}
H(\mappingFunctionVector\given\inducingVector) = \half \log \det{\conditionalCovariance}.
\shortlong{$}{\]}
%
From an information theoretic perspective, the conditional entropy
represents the amount of additional information we need to specify the
distribution of $\mappingFunctionVector$ given that we already have
the inducing vector, $\inducingVector$. This required additional
information can be manipulated by changing the relationship between
$\inducingVector$ and $\mappingFunctionVector$. The inducing points
themselves are parameterised by
$\kernelMatrix_{\mappingFunctionVector\inducingVector}$ and
$\kernelMatrix_{\inducingVector\inducingVector}$ which introduce a new
set of \emph{variational} parameters to the model. Those parameters
typically include the inducing input locations and any parameters of
the covariance function itself. We can be quite creative in how we
specify these relationships, for example, placing the inducing
variables in separate domain was suggested by
\citet{Alvarez:efficient10}. By minimizing this term we improve the
quality of the bound.

Computing this entropy requires $\bigO(\numData^3)$ computations due
to the log determinant, but the log determinant of the matrix is upper
bounded by its trace, this is trivially true because log determinant
is the sum of the \emph{log} eigenvalues of $\conditionalCovariance$ whereas the trace is the sum of the eigenvalues directly. The logarithm is
upper bounded by the linear function. So it is a sufficient condition
for $\trace{\conditionalCovariance}$ to be small, to minimize the
conditional entropy of the inducing point relation. The trace of a
covariance is sometimes referred to as the \emph{total variance} of
the distribution. We therefore call this bound on the conditional
entropy the total conditional variance (TCV).

The TCV of the density,
$p(\mappingFunctionVector|\inducingVector)$,
$\trace{\conditionalCovariance}$, appears in in
\refeq{eq:gp_conditional} and remains throughout \refeq{eq:titsias}
and \refeq{eq:svigp}. It controls conditional entropy and the
information in the data and upper bounds the additional information
that is contained in the data that is not represented by
$\inducingVector$. When this term is small, we expect the
approximation to work well. Indeed, the two extrema of the variational
compression behaviour are given when the TCV term is zero: either the
noise variance is very large scaling out the term or the TCV is zero
because the conditional entropy
$H(\mappingFunctionVector|\inducingVector)$ is zero.

The TCV ensures that the inducing variables take up appropriate
positions when manipulating $\inducingInputMatrix$ to tighten the
variational bound. It also plays a key role in the formulation for of the variational bound we now present for deep GPs.

\section{Variational Compression in Deep GPs}
\label{sec:deep}
Deep \shortlong{GPs}{Gaussian processes} are models of the form 
\begin{equation}
\dataVector = \hiddenVector_\ell(\hiddenVector_{\numLayers-1}(...\hiddenVector_1(\inputVector))) + \noiseVector,
\end{equation}
where we then make an assumption that each of the hidden (potentially multivariate) functions $\hiddenVector_i$ is given by a \shortlong{GP}{Gaussian process}. For a fixed set of inputs $\inputMatrix$ and a series of observed responses $\dataVector$, and taking the vector $\hiddenVector_i$ to contain the vector of variables representing function values in the $i^\text{th}$ layer, the joint probability can be written
\begin{equation}
  p(\dataVector, \{\hiddenVector_i\}_{i=1}^{\numLayers}\given\inputMatrix) = p(\dataVector\given\hiddenVector_{\numLayers-1})\prod_{i=2}^{\numLayers} p(\hiddenVector_i\given\hiddenVector_{i-1}) p(\hiddenVector_1\given\inputMatrix)
\label{eq:deep_structure}
\end{equation}
with
\shortlong{$\hiddenVector_1\given\inputVector  \sim \gaussianSamp{\zerosVector}{\kernelMatrix_{\hiddenVector_{1}\hiddenVector_1}+\sigma^2_1\eye}$,
$\hiddenVector_i\given\hiddenVector_{i-1}   \sim \gaussianSamp{\zerosVector}{\kernelMatrix_{\hiddenVector_{i}\hiddenVector_i}+\sigma^2_i\eye}$ and
$\dataVector\given\hiddenVector_{\numLayers-1}  \sim \gaussianSamp{\zerosVector}{\kernelMatrix_{\mappingFunctionVector_{\numLayers}\mappingFunctionVector_{\numLayers}}+\sigma^2_\numLayers\eye}.$
}{
\begin{align*}
\hiddenVector_1\given\inputVector & \sim \gaussianSamp{\zerosVector}{\kernelMatrix_{\hiddenVector_{1}\hiddenVector_1}+\sigma^2_1\eye}, \\
\hiddenVector_i\given\hiddenVector_{i-1} &  \sim \gaussianSamp{\zerosVector}{\kernelMatrix_{\hiddenVector_{i}\hiddenVector_i}+\sigma^2_i\eye},\\
\dataVector\given\hiddenVector_{\numLayers-1} & \sim \gaussianSamp{\zerosVector}{\kernelMatrix_{\mappingFunctionVector_{\numLayers}\mappingFunctionVector_{\numLayers}}+\sigma^2_\numLayers\eye}.
\end{align*}
}
Inference over the latent variables $\hiddenVector$ is very
challenging. Not only will computations scale with the usual
$\bigO(\numData^3)$ rule for GPs, but the hidden layer variables are
also dependent {\em between} layers. We turn to approximate Gaussian
processes based on variational compression. Then to deal with the
dependence between layers we will To deal with the dependence between
layers we use the VC trick again. The result is a tractable bound on
the marginal likelihood of a deep GP which is scalable and
interpretable.

\subsection{Augmenting Each Layer}

As per the original formulation \citep{Damianou:deepgp13}, we assume
that the function at each layer includes some independent Gaussian
noise with variance $\sigma^2_i$, and we augment that layer with a
set of inducing variables $\inducingVector_i$ in the same way as a for
a single-layer GP model. Within each layer, then, we can apply
variational compression to achieve a bound on the conditional
probability, as per equation \refeq{eq:gp_conditional}. Substituting
this result in to the structure \refeq{eq:deep_structure} results in
{\shortlong{\small}{}
\begin{align}
  p(\dataVector, \{\hiddenVector_i\}_{i=1}^{\numLayers-1}\given\{\inducingVector_i\}_{i=1}^{\numLayers}, \inputMatrix) &\geq 
  \tilde p(\dataVector\given\inducingVector_{\numLayers}, \hiddenVector_{\numLayers-1})\nonumber \\ & \times \prod_{i=2}^{\numLayers-1} \tilde p(\hiddenVector_i\given\inducingVector_i,\hiddenVector_{i-1}) \tilde p(\hiddenVector_1\given\inducingVector_i,\inputMatrix) \nonumber \\
  & \times
  \exp\left(\sum_{i=1}^\numLayers -\frac{1}{2\sigma^2_i}\trace{\conditionalCovariance_{i}}\right),
  \label{eq:deep_structure}
\end{align}
}
where we have omitted dependence on $\inducingInputMatrix$ variables
for clarity, and have defined
\shortlong{$}{\[}
\tilde p(\hiddenVector_i\given\inducingVector_i,\hiddenVector_{i-1})
= \gaussianDist{\hiddenVector_i}{\kernelMatrix_{\hiddenVector_{i}\inducingVector_{i}}\kernelMatrix_{\inducingVector_i\inducingVector_i}^{-1}\inducingVector_i}{\sigma^2_i\eye}.
\shortlong{$}{\]}
The conditional covariance matrices have been indexed by layer, so
that
\shortlong{$}{\[}
\conditionalCovariance_{i} =
\kernelMatrix_{\hiddenVector_i\hiddenVector_i} -
\kernelMatrix_{\hiddenVector_i\inducingVector_i}\kernelMatrix_{\inducingVector_i
  \inducingVector_i}^{-1}\kernelMatrix_{\inducingVector_i
  \hiddenVector_i}.
\shortlong{$}{\]}
\begin{figure}
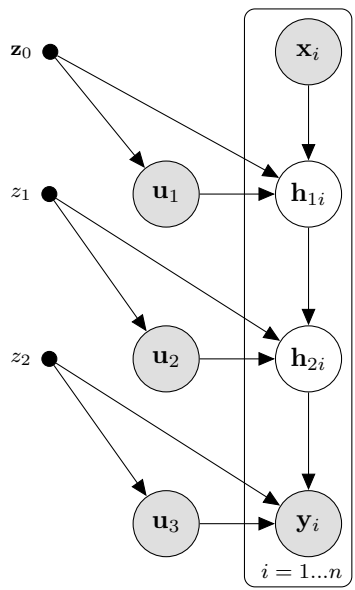

  \begin{center}
    \tikz{%
      \node[obs] (x) {$\mathbf \inputScalar_i$};
      \node[latent, below=1 of x] (h1) {$\hiddenVector_{1i}$};
      \node[obs, left=1 of h1] (u1) {$\inducingVector_{1}$};
      \node[const, left=2.88 of x, label=left:$\inducingInputVector_0$] (z0) {};
      \edge[] {z0}{u1};
      \edge[] x {h1};
      \edge[] {u1} {h1};
      \edge[] {z0} {h1};
      %
      \node[const, left=1 of u1, label=left:$\inducingInputScalar_1$] (z1) {};
      \node[latent, below=1.3 of h1] (h2) {$\hiddenVector_{2i}$};
      \node[obs, left=1 of h2] (u2) {$\inducingVector_{2}$};
      \edge[] {z1}{u2};
      \edge[] {h1} {h2};
      \edge[] {u2} {h2};
      \edge[] {z1} {h2};
      %
      \node[const, left=1 of u2, label=left:$\inducingInputScalar_2$] (z2) {};
      \node[obs, below=1.3 of h2] (y) {$\dataVector_{i}$};
      \node[obs, left=1 of y] (u3) {$\inducingVector_{3}$};
      \edge[] {z2}{u3};
      \edge[] {h2} {y};
      \edge[] {u3} {y};
      \edge[] {z2} {y};
      %
      \plate {yhx} {(y)(h2)(h1)(x)} {\,\,$i=1...\numData$};
    }
  \end{center}
  \caption{\small A graphical model representation of the deep GP structure
    with inducing variables. Nodes arranged horizontally form part of
    the same layer. we've added the index over $i$ to highlight the
    fact that conditioned on $\inducingVector_i$ we obtain
    independence over the data points.\label{fig:deep_depend}}
\end{figure}
If we ignore the TCV terms, then the remaining factors in
\refeq{eq:deep_structure} describe a joint probability density over
$\dataVector$ and $\mappingFunctionVector$ conditioned on
$\inducingVector$. \Reffig{fig:deep_depend} shows the resulting
graphical model associated with this distribution. 
The plate notation indicates that the terms
factorize across data points, this property will be preserved
when we apply variational compression through the layers.


Here, our work diverges from that of \citet{Damianou:deepgp13}. In
their work, the inducing variables $\inducingVector$ are marginalised
(similarly to equation \refeq{eq:titsias}), and a variational
distribution $q(\hiddenVector)$ is introduced for the hidden
variables. This has two consequences: marginalising $\inducingVector$
re-introduces dependencies between the hidden variables within a
layer, resulting in an objective function which cannot be written as a
sum of independent data terms. Secondly, the size of the optimization
problem grows with $\numData$, as distributions representing latent
variables have to be introduced, corresponding to each datum in each
of the hidden layers. In this alternative formulation we will seek to
retain the factorization across the data points allowing us to
consider stochastic variational inference approaches to the model.

\shortlong{\paragraph{The First Layer}}{\subsection{The First Layer}}
We now apply the variational compression approach to marginalise the variables associated with the first \shortlong{GP}{Gaussian process}. Although to ensure the cancellation we now assume that
\begin{equation}
  q(\hiddenVector_1\given\inducingVector_1) = \tilde p(\hiddenVector_1\given\inducingVector_1, \inputMatrix),
  \label{eq:layer1_partial}
\end{equation}
and form a bound on the conditional distribution. Taking the relevant
terms from equation \refeq{eq:deep_structure}, we obtain
\begin{align}
  \log p(\hiddenVector_2\given\inducingVector_1, \inducingVector_2) \geq &\tilde p(\hiddenVector_1\given\inducingVector_1) \expSamp{\log \tilde{p}(\hiddenVector_2\given\hiddenVector_1, \inducingVector_2)} \nonumber \\
  & -\expSamp{\frac{1}{2\sigma^2_2}\trace{\boldsymbol{\Sigma_1}}}
  -\frac{1}{2\sigma_1^2}\trace{\conditionalCovariance_0}.
  \label{eq:layer1_conditional}
\end{align}

The second stage of variational compression is to marginalize
$\inducingVector_1$. Since the above expression is not conjugate to
the Gaussian prior $p(\inducingVector_1)$, we must 
introduce variational parameters
$q(\inducingVector_1) = \gaussianDist{\inducingVector_1}{\mathbf
  m_1}{\mathbf S_1}$. Along with \refeq{eq:layer1_partial}, we now
have:
\begin{equation}
q(\hiddenVector_1) = \int \tilde p(\hiddenVector_1\given\inducingVector_1)q(\inducingVector_1)\text{d}\inducingVector_1, 
\end{equation}
which is a straightforward Gaussian integral. 

Although our approximation to the latent space is {\em not} factorized
across the data dimension $\numData$, we are still able to construct
an algorithm that depends on the data points independently; in
practice, we need only ever compute the diagonal parts of the
covariance in $q(\hiddenVector)$ at each layer.

Using this definition of $q(\mappingFunctionVector_1)$ with equation \eqref{eq:layer1_conditional} allows us to variationally marginalize $\inducingVector_1$:
\begin{align}
  \log & p(\hiddenVector_2\given\inputMatrix, \inducingVector_2) \geq \nonumber \\
&-\frac{1}{\sigma_1^2} \trace{\conditionalCovariance_0}
-\frac{1}{\sigma_2^2}\expDist{\trace{\conditionalCovariance_1}}{q(\hiddenVector_1)} \nonumber \\
&- \KL{q(\inducingVector_1)}{p(\inducingVector_1)} + \nonumber \\
& \expDist{\log \gaussianDist {\hiddenVector_2}{\kernelMatrix_{\hiddenVector_2, \inducingVector_2}\kernelMatrix_{\inducingVector_2\inducingVector_2}^{-1}\inducingVector_2}{\sigma^2_2\eye)}}{q(\hiddenVector_1)}.
\label{eq:first_layer_partly}
\end{align}

The expectations under $q(\hiddenVector_1)$ involve the covariance
function in the same way as
\citet{Titsias:bayesGPLVM10,Damianou:deepgp13}. The required
quantities are
\shortlong{$\psi_i = \expDist{\trace{\kernelMatrix_{\hiddenVector_i\hiddenVector_i}}}{q(\hiddenVector_{i-1})}$,
${\boldsymbol \Psi}_i = \expDist{\kernelMatrix_{\hiddenVector_i\inducingVector_i}}{q(\hiddenVector_{i-1})}$ and
${\boldsymbol \Phi}_i = \expDist{\kernelMatrix_{\inducingVector_i\hiddenVector_i}\kernelMatrix_{\hiddenVector_i\inducingVector_i}}{q(\hiddenVector_{i-1})}.$
}{\begin{align*}
  \psi_i &= \expDist{\trace{\kernelMatrix_{\hiddenVector_i\hiddenVector_i}}}{q(\hiddenVector_{i-1})}\\
  {\boldsymbol \Psi}_i &= \expDist{\kernelMatrix_{\hiddenVector_i\inducingVector_i}}{q(\hiddenVector_{i-1})}\\
  {\boldsymbol \Phi}_i &= \expDist{\kernelMatrix_{\inducingVector_i\hiddenVector_i}\kernelMatrix_{\hiddenVector_i\inducingVector_i}}{q(\hiddenVector_{i-1})}.
\end{align*}}
which can be computed analytically for some popular choices of
covariance function including the exponentiated quadratic,
\shortlong{$}{\[}
\kernelScalar(\inputVector, \inputVector^\prime) = \alpha
\exp\left(\frac{1}{\lengthScale^2}\ltwoNorm{\inputVector-\inputVector^\prime}^2\right),
\shortlong{$}{\]}
and linear forms,
\shortlong{$}{\[}
\kernelScalar(\inputVector, \inputVector^\prime) =
\inputVector^\top \inputVector.
\shortlong{$}{\]} 
With these definitions, it is possible to substitute into \refeq{eq:first_layer_partly} and re-arrange to give
{\shortlong{\small}{}
\begin{equation}
  \begin{split}
    \log p(\hiddenVector_2\given\inputMatrix, \inducingVector_2) & \geq 
    -\frac{1}{\sigma_1^2} \trace{\conditionalCovariance_0} 
    %
    - \frac{1}{\sigma_2^2} \expDist{\trace{\conditionalCovariance_1}}{q(\hiddenVector_1)}\\
    &- \KL{q(\inducingVector_1)}{p(\inducingVector_1)}\\
     - \frac{1}{\sigma_2^2} & \trace{({\boldsymbol \Phi}_2 - {\boldsymbol \Psi}_2^\top{\boldsymbol \Psi}_2) \kernelMatrix_{\inducingVector_2 \inducingVector_2}^{-1}\inducingVector_2\inducingVector_2^\top\kernelMatrix_{\inducingVector_2\inducingVector_2}^{-1}}\\
    &+ \log \gaussianDist{\hiddenVector_2}{{\boldsymbol \Psi}_2\kernelMatrix_{\inducingVector_2\inducingVector_2}^{-1}\inducingVector_2}{\sigma^2_2\eye}\\
    \label{eq:first_layer_done}
  \end{split}
\end{equation}}
where we have completed the square to ensure that the
$\hiddenVector_2$ appears in a normalised Gaussian distribution. We
are left with a conditional expression for the second layer, where the
information from the layer above has been propagated variationally, as
well as a series of terms that ensure that the expression remains a
variational bound.

\shortlong{\paragraph{Subsequent Layers}}{\subsection{Subsequent Layers}}

Having arrived at an expression for the second layer, with the first
layer marginalized, we are in a position to apply variational
compression again to marginalize layer 2. The procedure follows much
the same pattern as for the first layer.
The
result is a bound on $p(\hiddenVector_3\given\inducingVector_3)$, with
some additional terms, where the relation
$\hiddenVector_3\given\inducingVector_3$ is again a normal
distribution.

Recursively applying this formulation through an arbitrarily deep
network results in the following bound:
{\shortlong{\small}{}
\begin{align}
\log p(\dataVector\given\inputMatrix )  \geq &
-\frac{1}{\sigma_1^2} \trace{\conditionalCovariance_1}
-\sum_{i=2}^\numLayers \frac{1}{2\sigma_i^2} \left(\psi_{i}
  - \trace{{\boldsymbol \Phi}_{i}\kernelMatrix_{\inducingVector_{i} \inducingVector_{i}}^{-1}}\right) \nonumber \\
& - \sum_{i=1}^{\numLayers}\KL{q(\inducingVector_i)}{p(\inducingVector_i)} \nonumber \\
 - \sum_{i=2}^{\numLayers}\frac{1}{2\sigma^2_{i}} & \trace{({\boldsymbol
    \Phi}_i - {\boldsymbol \Psi}_i^\top{\boldsymbol \Psi}_i)
  \kernelMatrix_{\inducingVector_{i}
    \inducingVector_{i}}^{-1}
  \expDist{\inducingVector_{i}\inducingVector_{i}^\top}{q(\inducingVector_{i})}\kernelMatrix_{\inducingVector_{i}\inducingVector_{i}}^{-1}} \nonumber \\
& + \log \gaussianDist{\dataVector}{{\boldsymbol
    \Psi}_{\numLayers}\kernelMatrix_{\inducingVector_{\numLayers}
    \inducingVector_{\numLayers}}^{-1}{\mathbf
    m}_\numLayers}{\sigma^2_\numLayers\eye}
\label{eq:deep_bound}
\end{align}}
This expression constitutes the main contribution of this work. We now have a
bound on the marginal likelihood, parameterised by variational distributions
$q(\inducingVector_i)$ at each layer, along with inducing inputs for each
layer $\inducingInputMatrix_i$ and covariance parameters of the kernel at
each layer. Importantly, each part of the bound can be written as a sum of
$\numData$ terms, with each term depending on only one datum. This allows
straightforward inference in the model using parallelized or stochastic
methods, without having to deal with large numbers of latent variables,
$\hiddenVector_i$. 

All the latent variables $\{\hiddenVector_i\}_{i=1}^\ell$ have now
been marginalised using our variational compression scheme. From the
previous discussion, we know that the approximation will be reasonable
if they are highly correlated with the inducing points
$\inducingVector_i$. These correlations are reduced if we inject a
larger amount of noise at each hidden layer: if the variance variables
$\sigma^2_i$ are large, then our approximation will fail.

\shortlong{\paragraph{Examining the Bound}}{\subsection{Examining the Bound}}

Our bound on the marginal likelihood has a single `likelihood' expression
(equation \refeq{eq:deep_bound}, last line), preceded by a series of
`regularizing' terms. We first examine the likelihood part, which has a similar
form to a neural network. The term is Gaussian with mean given by a linear
combination of the columns of ${\boldsymbol \Psi}_\numLayers$, with weights, $\kernelMatrix_{\inducingVector_\numLayers\inducingVector_\numLayers} \mathbf m_\numLayers$, which are given by the mean of the variational distribution $q(\inducingVector_\numLayers)$ weighted by its prior covariance. Information from previous layers feeds in through the matrix,
\shortlong{$}{\[}
{\boldsymbol \Psi}_{\numLayers} = \expDist{\kernelMatrix_{\mappingFunctionVector_\numLayers\inducingVector_\numLayers}}{q(\mappingFunctionVector_{\numLayers-1})}.
\shortlong{$}{\]}
We can examine the $j$th column of this matrix for the $i$th input in the case of the exponentiated quadratic covariance,
\shortlong{$}{\[}
\psi^{\numLayers}_{i, j} = \expDist{\alpha\exp\left(-\frac{\ltwoNorm{\mappingFunctionVector_{\numLayers-1} - \inducingInputVector_{\numLayers-1}}^2}{2\lengthScale^2}\right)}{q(\mappingFunctionVector_{\numLayers-1})}.
\shortlong{$}{\]}
We can compare this with a radial basis function network. In deep
versions of those models the $i$th layer the basis functions would be
centred on particular values, $\mathbf{c}_{i}$. Those centres are
functionally equivalent to our inducing inputs
$\inducingInputVector_{\numLayers-1}$. However, in our model they are
variational parameters, rather than model parameters. This is an
important difference because we can increase the number of them at any
time without risk of overfitting. If we \emph{do} use more centres we
can only reduce the conditional entropy
$H(\mappingFunctionVector_i|\inducingVector_i)$ thereby improving the
quality of the variational bound. In each layer, we can adjust the
weights
$\kernelMatrix_{\inducingVector_\numLayers\inducingVector_\numLayers}
{\mathbf m}_\numLayers$ through varying the mean of the variational
distribution over the inducing variables. If we were to use other
covariance functions we would obtain similar neural network-like
models but with different activation functions.

A significant difference to a neural network approach is that we are
propagating \emph{distributions} through each layer of the network
rather than just point values. In each layer we must compute
$q(\hiddenVector_i) = \int \tilde
p(\hiddenVector_i\given\inducingVector_i)q(\inducingVector_i)
\text{d}\inducingVector_i$, and propagate this to the next layer,
where it is used to take expectations of the kernel function to pass
into the subsequent layers. The form of our variational compression
means that these distributions are Gaussian, so we are propagating
both a mean and a variance through the model at every layer. On
computing the derivative of any part of the approximation, we must do
a feed-forward pass of these Gaussian messages, followed by a
backpropagation step using the chain-rule.

Our bound contains the trace of the conditional matrix (now in
expectation), which we have studied in section
\ref{par:trace_term}. It also contains the usual KL divergence term
from the prior. The final `regularization' term (third line of
\refeq{eq:deep_bound}) has interesting regularization properties: it
contains the variance of the GP function at each layer, under our
approximation. To see this, if we are given the values of a GP
function $\inducingVector$ at $\inducingInputMatrix$, the mean of the
prediction for the points $\mappingFunctionVector$ at $\inputMatrix$
is given by $\langle \mappingFunctionVector \rangle =
\kernelMatrix_{\mappingFunctionVector\inducingVector}
\kernelMatrix_{\inducingVector\inducingVector}^{-1}\inducingVector$.
The sum of the variance of this vector is then
$\trace{\kernelMatrix_{ \mappingFunctionVector\inducingVector
}\kernelMatrix_{\inducingVector\inducingVector
}^{-1}\inducingVector\inducingVector^\top
\kernelMatrix_{\inducingVector\inducingVector
}^{-1}\kernelMatrix_{\inducingVector\mappingFunctionVector}}$, and the
`regularization' term is the variance of this under
$q(\hiddenVector_{i-1})$, in expectation under
$q(\inducingVector)$. The term summarizes how the values of the
function will change as the input to the function changes: a measure
of the local `wiggliness' of the function under the variational input
distribution.

We refer to these two regularization terms as the `compression term' and  the
`propagation term'. In \reffig{fig:messages}, we illustrate the forward
passing of the Gaussian distributions (according to our likelihood computation)
for different scenarios, and the incurred penalty terms. In summary, the
regularization terms encourage approximations that can be well represented by
the limited representative power of $q(\inducingVector)$. 

\begin{figure*}
  \setlength{\figurewidth}{18cm}
  \setlength{\figureheight}{12cm}
  \centering\includegraphics[width=\textwidth]{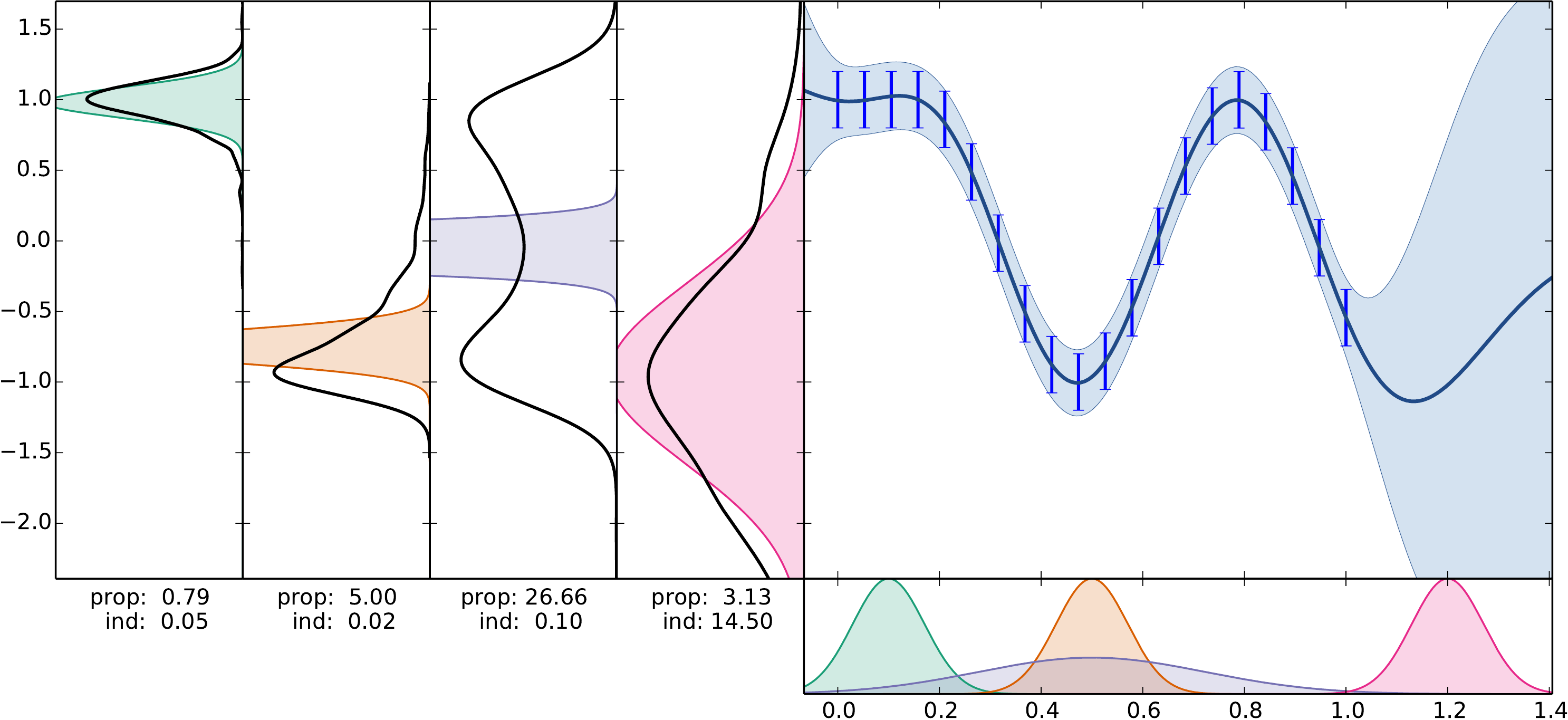}
  \caption{\small \label{fig:messages} Forward propagation of Gaussian messages
    through a layer of a deep GP. Bottom right: four colour-coded
    Gaussian distributions to be passed through the GP layer. Top-right:
    a \shortlong{GP}{Gaussian process}, represented by a finite series of inducing
    variables $\inducingVector$, depicted by vertical red bars. Left:
    the responses of the GP to the input distributions, with Gaussian
    approximation shown in colours matching bottom right, and the ground
    truth represented by a solid line, computed by Monte Carlo. Below
    each approximation we give the two penalty terms associated with
    propagating this distribution. The propagation term (top) is large
    when the function is locally highly nonlinear, and the compression
    term (below) is large when the input is far from the inducing
  variables.}
\end{figure*}

\section{Experiments}
\label{sec:experiments}

We present three applications of the variationally compressed deep GP method.
In each case, we used simple gradient based optimization to maximize the bound
on the marginal likelihood with respect to the covariance function parameters,
the inducing input positions $Z$ in each layer, and the variational parameters
of $q(\inducingVector)$ in each layer, $m_i, {\mathbf S}_i$. To maintain
positive-definiteness of the covariance matrices, we employed a
lower-triangular factorized representation, ${\mathbf S}_i = {\mathbf L}_i{\mathbf L}_i^\top$.

To exploit the factorised nature of the objective function, our python
implementation uses MPI to parallelize computation across several cores of a
desktop machine. Parallelism is across chunks of data, since equation \ref{eq:deepBound}
can be written as a sum of data dependent terms. Parallelization to larger
systems is straightforward, and the objective lends itself to stochastic
optimization. 

\shortlong{\paragraph{Step Function Data}}{\subsection{Step Function Data}}
In \citet{Rasmussen:book06}, a simple example is presented which is
challenging for standard GP regression: data representing a noisy step
function as in Figure \ref{fig:step}. The GP with exponentiated
quadratic covariance is unable to satisfactorily fit to the data, as
the top plot in Figure \ref{fig:step} shows. \citet{Rasmussen:book06}
propose using a neural-network based covariance function, which is
able to model the data somewhat better.  \citet{Calandra:manifold}
propose a two layer model where the data are transformed by some
deterministic function before being passed into a GP, which appears to
perform somewhat better, though this is arguably still a kernel
selection problem, with parameters of the deterministic function being
incorporated into the kernel.
\begin{figure}
  \setlength{\figurewidth}{1.1\linewidth}
  \shortlong{
  \setlength{\figureheight}{2.5cm}
    }{
  \setlength{\figureheight}{4.5cm}}
%
%
%
%
\begin{tikzpicture}

\begin{axis}[
xmin=-1, xmax=2,
ymin=-0.3, ymax=1.3,
axis on top,
width=\figurewidth,
height=\figureheight,
xticklabels=\empty
]
\addplot [line width=0.68pt, black, mark=x, mark size=4, only marks]
coordinates {
(0,0.00480755177651519)
(0.0344827586206896,0.00954424471134621)
(0.0689655172413793,-0.0254630580388837)
(0.103448275862069,0.0310929817801382)
(0.137931034482759,0.0349264958169793)
(0.172413793103448,-0.0270507333419196)
(0.206896551724138,-0.0217430031252698)
(0.241379310344828,-0.0250896097066597)
(0.275862068965517,0.0249386823582826)
(0.310344827586207,-0.0466705060812323)
(0.344827586206897,0.00260954138435669)
(0.379310344827586,0.0298698319183808)
(0.413793103448276,0.000761584988140691)
(0.448275862068965,0.0242831143942044)
(0.482758620689655,-0.000715789737214675)
(0.517241379310345,0.986235145248351)
(0.551724137931034,1.00038983548784)
(0.586206896551724,1.00655462965101)
(0.620689655172414,0.931558620335998)
(0.655172413793103,1.00263469679235)
(0.689655172413793,0.989521577579133)
(0.724137931034483,1.01530874527295)
(0.758620689655172,1.01088003470569)
(0.793103448275862,1.03758984926622)
(0.827586206896552,1.01691420485221)
(0.862068965517241,0.996536913189708)
(0.896551724137931,1.01706143761017)
(0.931034482758621,0.996925127543056)
(0.96551724137931,0.979836909040802)
(1,0.99177048985369)

};
\addplot graphics [includegraphics cmd=\pgfimage,xmin=-1, xmax=2, ymin=-0.3, ymax=1.3] {\todiagrams 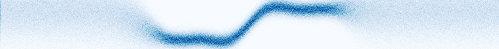};
\path [draw=black, fill opacity=0] (axis cs:13,1.3)--(axis cs:13,1.3);

\path [draw=black, fill opacity=0] (axis cs:2,13)--(axis cs:2,13);

\path [draw=black, fill opacity=0] (axis cs:13,-0.3)--(axis cs:13,-0.3);

\path [draw=black, fill opacity=0] (axis cs:-1,13)--(axis cs:-1,13);

\end{axis}

\end{tikzpicture}
%
%
%
%
\begin{tikzpicture}

\begin{axis}[
xmin=-1, xmax=2,
ymin=-0.3, ymax=1.3,
axis on top,
width=\figurewidth,
height=\figureheight,
xticklabels=\empty
]
\addplot [line width=0.68pt, black, mark=x, mark size=4, only marks]
coordinates {
(0,0.00480755177651519)
(0.0344827586206896,0.00954424471134621)
(0.0689655172413793,-0.0254630580388837)
(0.103448275862069,0.0310929817801382)
(0.137931034482759,0.0349264958169793)
(0.172413793103448,-0.0270507333419196)
(0.206896551724138,-0.0217430031252698)
(0.241379310344828,-0.0250896097066597)
(0.275862068965517,0.0249386823582826)
(0.310344827586207,-0.0466705060812323)
(0.344827586206897,0.00260954138435669)
(0.379310344827586,0.0298698319183808)
(0.413793103448276,0.000761584988140691)
(0.448275862068965,0.0242831143942044)
(0.482758620689655,-0.000715789737214675)
(0.517241379310345,0.986235145248351)
(0.551724137931034,1.00038983548784)
(0.586206896551724,1.00655462965101)
(0.620689655172414,0.931558620335998)
(0.655172413793103,1.00263469679235)
(0.689655172413793,0.989521577579133)
(0.724137931034483,1.01530874527295)
(0.758620689655172,1.01088003470569)
(0.793103448275862,1.03758984926622)
(0.827586206896552,1.01691420485221)
(0.862068965517241,0.996536913189708)
(0.896551724137931,1.01706143761017)
(0.931034482758621,0.996925127543056)
(0.96551724137931,0.979836909040802)
(1,0.99177048985369)

};
\addplot graphics [includegraphics cmd=\pgfimage,xmin=-1, xmax=2, ymin=-0.3, ymax=1.3] {\todiagrams 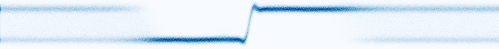};
\path [draw=black, fill opacity=0] (axis cs:13,1.3)--(axis cs:13,1.3);

\path [draw=black, fill opacity=0] (axis cs:2,13)--(axis cs:2,13);

\path [draw=black, fill opacity=0] (axis cs:13,-0.3)--(axis cs:13,-0.3);

\path [draw=black, fill opacity=0] (axis cs:-1,13)--(axis cs:-1,13);

\end{axis}

\end{tikzpicture}
%
%
%
%
\begin{tikzpicture}

\begin{axis}[
xmin=-1, xmax=2,
ymin=-0.3, ymax=1.3,
axis on top,
width=\figurewidth,
height=\figureheight
]
\addplot [line width=0.68pt, black, mark=x, mark size=4, only marks]
coordinates {
(0,0.00480755177651519)
(0.0344827586206896,0.00954424471134621)
(0.0689655172413793,-0.0254630580388837)
(0.103448275862069,0.0310929817801382)
(0.137931034482759,0.0349264958169793)
(0.172413793103448,-0.0270507333419196)
(0.206896551724138,-0.0217430031252698)
(0.241379310344828,-0.0250896097066597)
(0.275862068965517,0.0249386823582826)
(0.310344827586207,-0.0466705060812323)
(0.344827586206897,0.00260954138435669)
(0.379310344827586,0.0298698319183808)
(0.413793103448276,0.000761584988140691)
(0.448275862068965,0.0242831143942044)
(0.482758620689655,-0.000715789737214675)
(0.517241379310345,0.986235145248351)
(0.551724137931034,1.00038983548784)
(0.586206896551724,1.00655462965101)
(0.620689655172414,0.931558620335998)
(0.655172413793103,1.00263469679235)
(0.689655172413793,0.989521577579133)
(0.724137931034483,1.01530874527295)
(0.758620689655172,1.01088003470569)
(0.793103448275862,1.03758984926622)
(0.827586206896552,1.01691420485221)
(0.862068965517241,0.996536913189708)
(0.896551724137931,1.01706143761017)
(0.931034482758621,0.996925127543056)
(0.96551724137931,0.979836909040802)
(1,0.99177048985369)

};
\addplot graphics [includegraphics cmd=\pgfimage,xmin=-1, xmax=2, ymin=-0.3, ymax=1.3] {\todiagrams 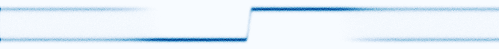};
\path [draw=black, fill opacity=0] (axis cs:13,1.3)--(axis cs:13,1.3);

\path [draw=black, fill opacity=0] (axis cs:2,13)--(axis cs:2,13);

\path [draw=black, fill opacity=0] (axis cs:13,-0.3)--(axis cs:13,-0.3);

\path [draw=black, fill opacity=0] (axis cs:-1,13)--(axis cs:-1,13);

\end{axis}

\end{tikzpicture}
  \vspace{-0.5cm}
  \caption{\small \label{fig:step}Regression on the noisy step-function data
    \citep{Rasmussen:book06}. Top: \shortlong{GP}{Gaussian process} regression,
    middle: deep GP with one hidden layer, bottom: deep GP with two
    hidden layers. In each plot, the posterior predictive density
    is shown in blue. }
\end{figure}
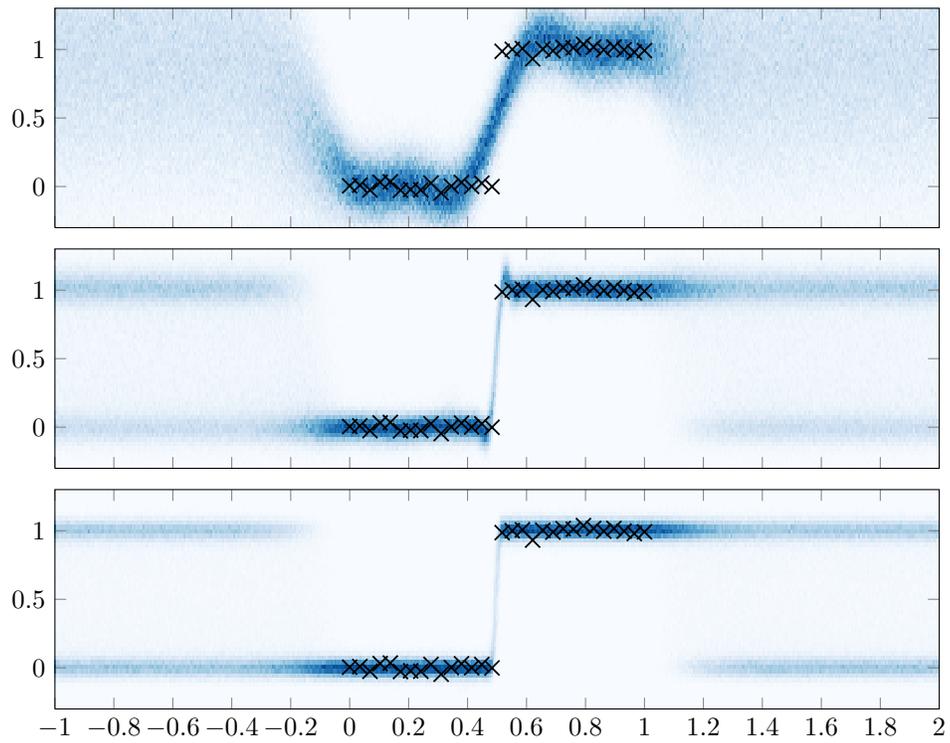

The idea of deep networks is to avoid this kind of kernel selection,
using layers of representation to infer features from the data. Can
deep GPs do away with the problem of having to select a kernel in a
Bayesian fashion? Figure \ref{fig:step} suggests that this might be
the case. Moving down the panels of the Figure, the depth of the model
is increased from 1 (a GP model) to 3. The GP model is unable to cope
with the non-smoothness of the function, provides an `overshot' mean
function and extrapolates poorly. The two layer deep GP has more
capability to fit the sharp change in the function though it still
overshoots a little, and the three layer deep model performs well,
with a sharp mean function at the step. Extrapolation from the data
has interesting behaviour, we note that despite the posterior appearing
bimodal, any sample drawn from the posterior will be a continuous
function, which will switch between the two modes with a lengthscale
of around 0.2.

\shortlong{\paragraph{Robot Wireless Data}}{\subsection{Robot Wireless Data}}
Our second data set consists of a robotics localization problem
\citep{Ferris:wifi07}. We are presented with the signal strengths of thirty
wireless access points located around a building, as detected by a
robot which is tracing a path through the corridors. The wireless
signal strengths fluctuate as a function of time as the robot moves
nearer of further from each access point. The underlying structure of
the signals must surely represent the position of the robot in the
building.

\begin{figure*}
\centering
\newlength\robotwidth
\setlength{\robotwidth}{\shortlong{0.2\textwidth}{0.4\textwidth}}
\begin{subfigure}[b]{\robotwidth}
  \includegraphics[width=\textwidth]{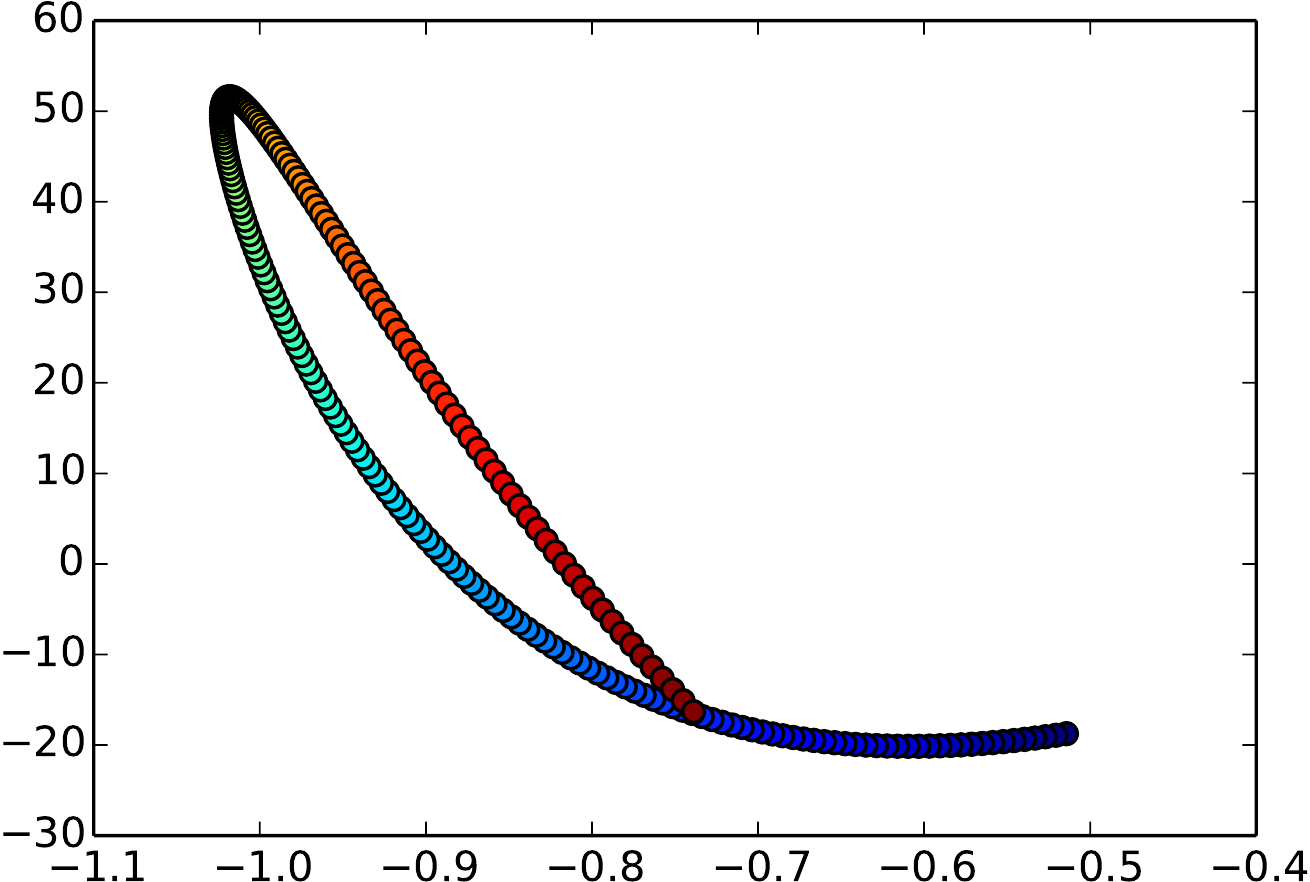}
  \caption{First hidden layer}
\end{subfigure}
\quad
\begin{subfigure}[b]{\robotwidth}
  \includegraphics[width=\textwidth]{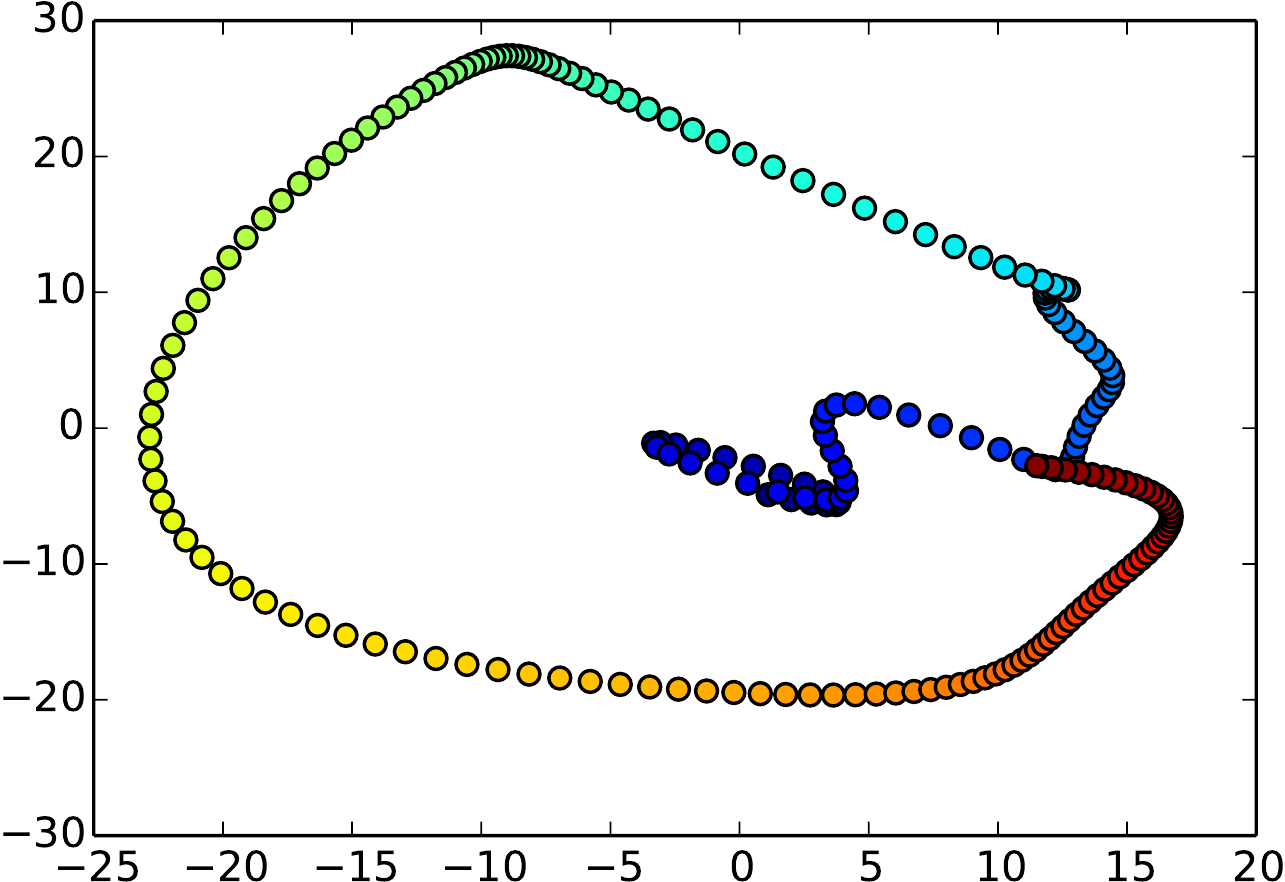}
  \caption{Second hidden layer}
\end{subfigure}
\quad
\begin{subfigure}[b]{\robotwidth}
  \includegraphics[width=\textwidth]{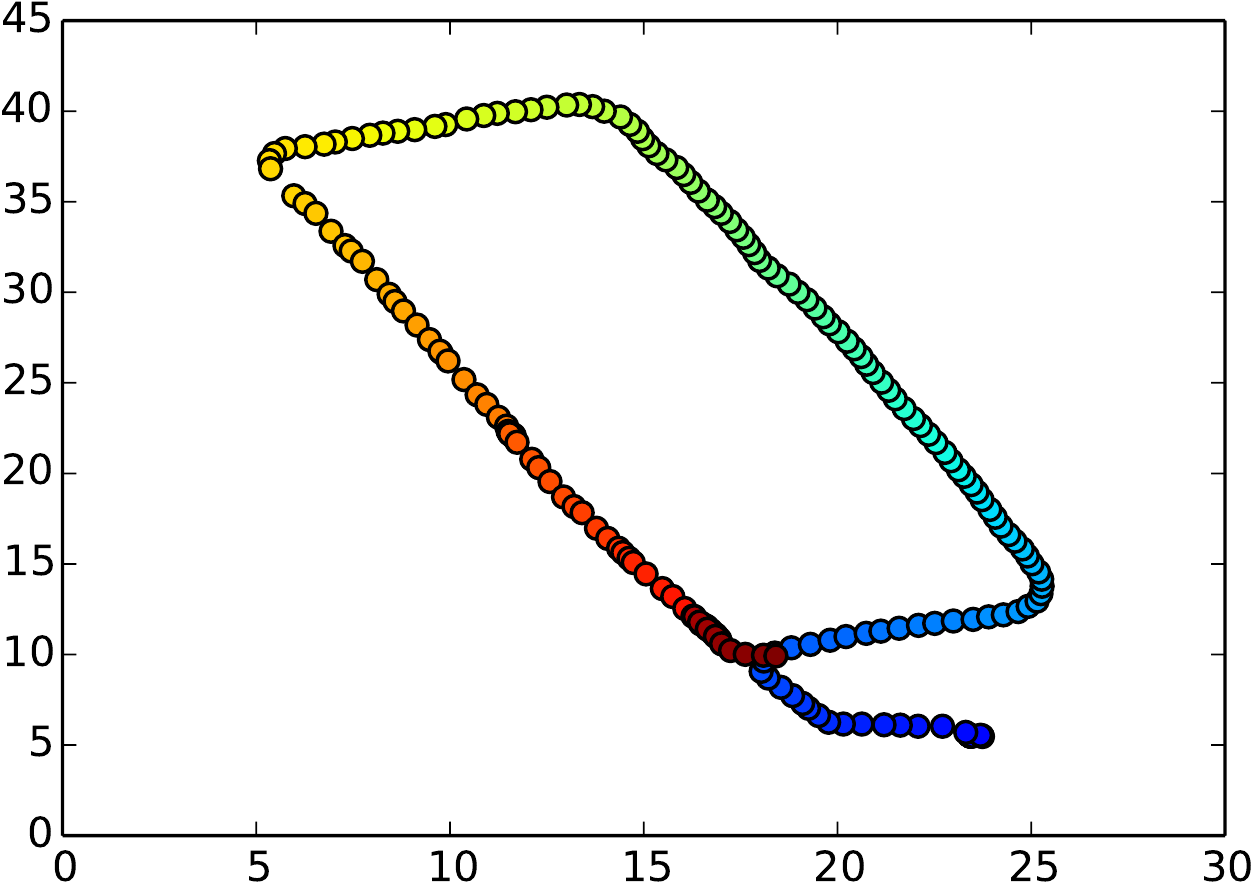}
  \caption{True robot path}
\end{subfigure}
\quad
\begin{subfigure}[b]{\robotwidth}
  \includegraphics[width=\textwidth]{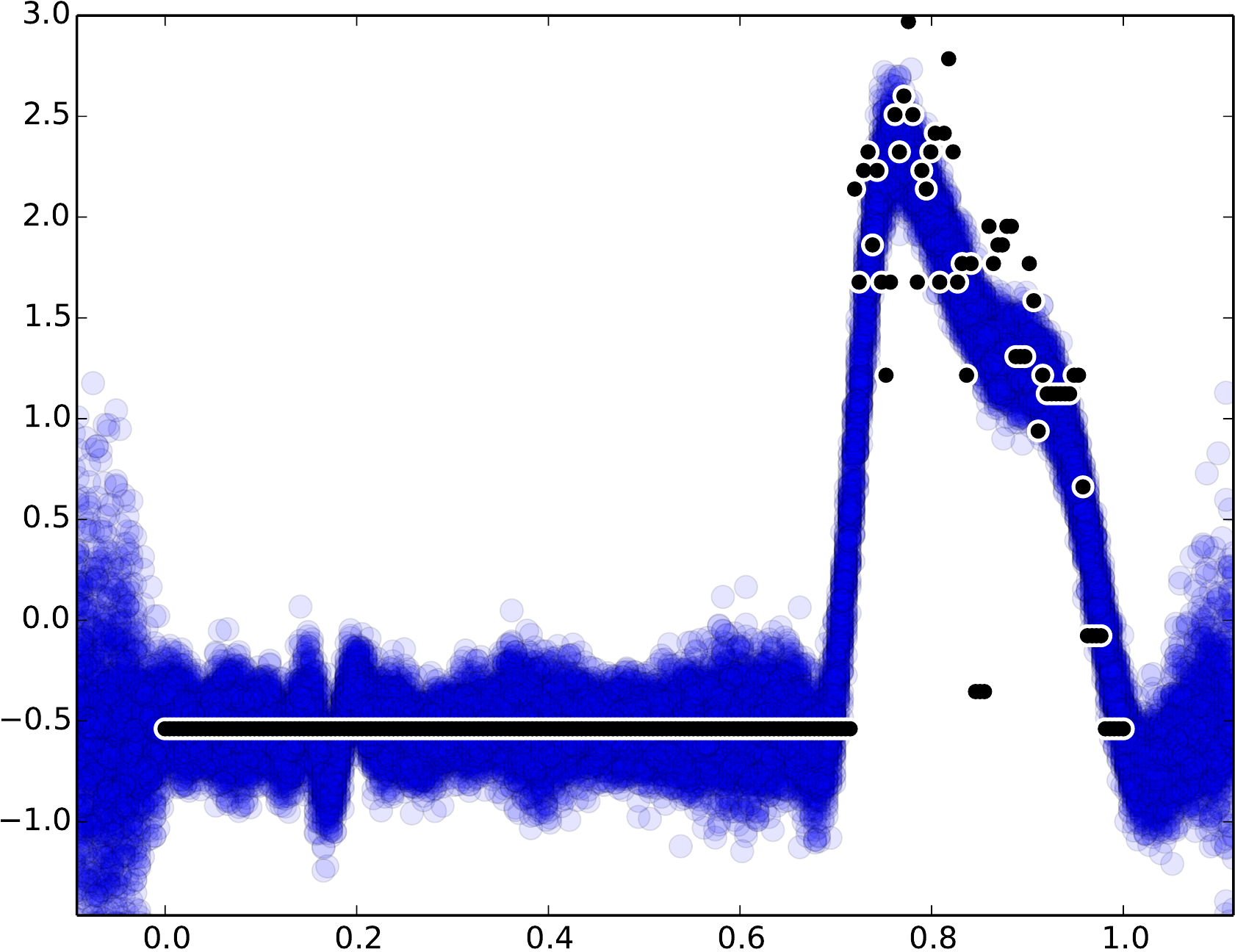}
  \caption{Typical signal and deep GP fit}
\end{subfigure}
\caption{\small \label{fig:robots} Latent representation of the robot
  wireless data. In each figure, time (the input to the deep model) is
  represented by color, and each circle represents one temporal data
  point. (a) the latent representation in the first layer.  (b) The
  latent representation in the second layer. (c) The ground truth: the
  known path of the robot. The latent representations qualitatively
  represent the topology (a) and finer structure (b) of the ground
  truth. (d) Shows a typical signal, with data represented by black
  points and the posterior density for this signal in blue.}
\end{figure*}

We build a 3 layer deep GP, in which the input layer was time, through
two hidden layers with 3 dimensions each, and finally mapping to the
30 dimensional signal strengths. After optimization of the
marginal-likelihood bound using L-BFGS-B \citep{Zhu:lbfgsb97}, the
learned structure of the model is as in Figure \ref{fig:robots}.

\Reffig{fig:robots} (c) shows the true path of the robots, which completes
a rectangular loop around the building, with a 'tail' of data at the start. In
the fist hidden layer (\reffig{fig:robots}(a)), the models learns the
topology of the structure: a loop closes at the correct point. In the
subsequent layer, the representation contains `corners', representing structure
in the data where correlations in the signal strength must vary rapidly or
slowly. This hierarchically structured learning of features is a key feature
of the deep GP model which is not possible using a standard GP. 

The final frame of \reffig{fig:robots} shows on of the thirty
signals used.  Two interesting features are prominent: the deep GP is
insensitive to outliers (some occur at around $t=0.85$), and the deep GP
predicts structure where none is present (around $t=0.2$). Both of these
effects are a result of the strong structural prior imposed by the
deep GP model: output variables are expected to covary in a consistent
way, and structure in some part of the data is imposed on other parts
also.

In both these cases, the strong structural prior is reasonable. At
around $t=0.2$ the WiFi drops out because the device only retains the
largest signals it can measure. However, the true underlying signal is
still likely to covary with the other measurements in a fashion as
predicted by the model. The outlying data are also explained well by
the deep GP.

The variational approximation to the posterior is unimodal,
and the deep GP prior contains many possible modes which can be
constructed by symmetrical and rotational arguments. Re-running the
experiment shown in \reffig{fig:robots} will result in a different
representation each time. Selection of the `best' approximating
posterior is possible by comparing the bounds on the marginal
likelihood. We note that the majority of solutions found for this
problems qualitatively represent that presented in \reffig{fig:robots}. There
would also be scope to combine the variational approximations through mixture
distributions \citep{Lawrence:thesis00}.

\shortlong{\paragraph{Autoencoders}}{\subsection{Autoencoders}}
Building auto-encoding deep GPs is straightforward: we simple use the
same data as input and output to the model.

We build a two-layer deep GP, with the same data on the input and
output. That is, a single hidden representation with an `encoding'
layer and a `decoding' layer. Using the Frey faces dataset
\citep{Frey:mixtures98}, which has 784 dimensions (pixels) and
approximately 2,000 training points, we optimized the marginal
likelihood bound using L-BFGS-B \citep{Zhu:lbfgsb97}.

\begin{figure}
  \centering\includegraphics[width=0.8\columnwidth]{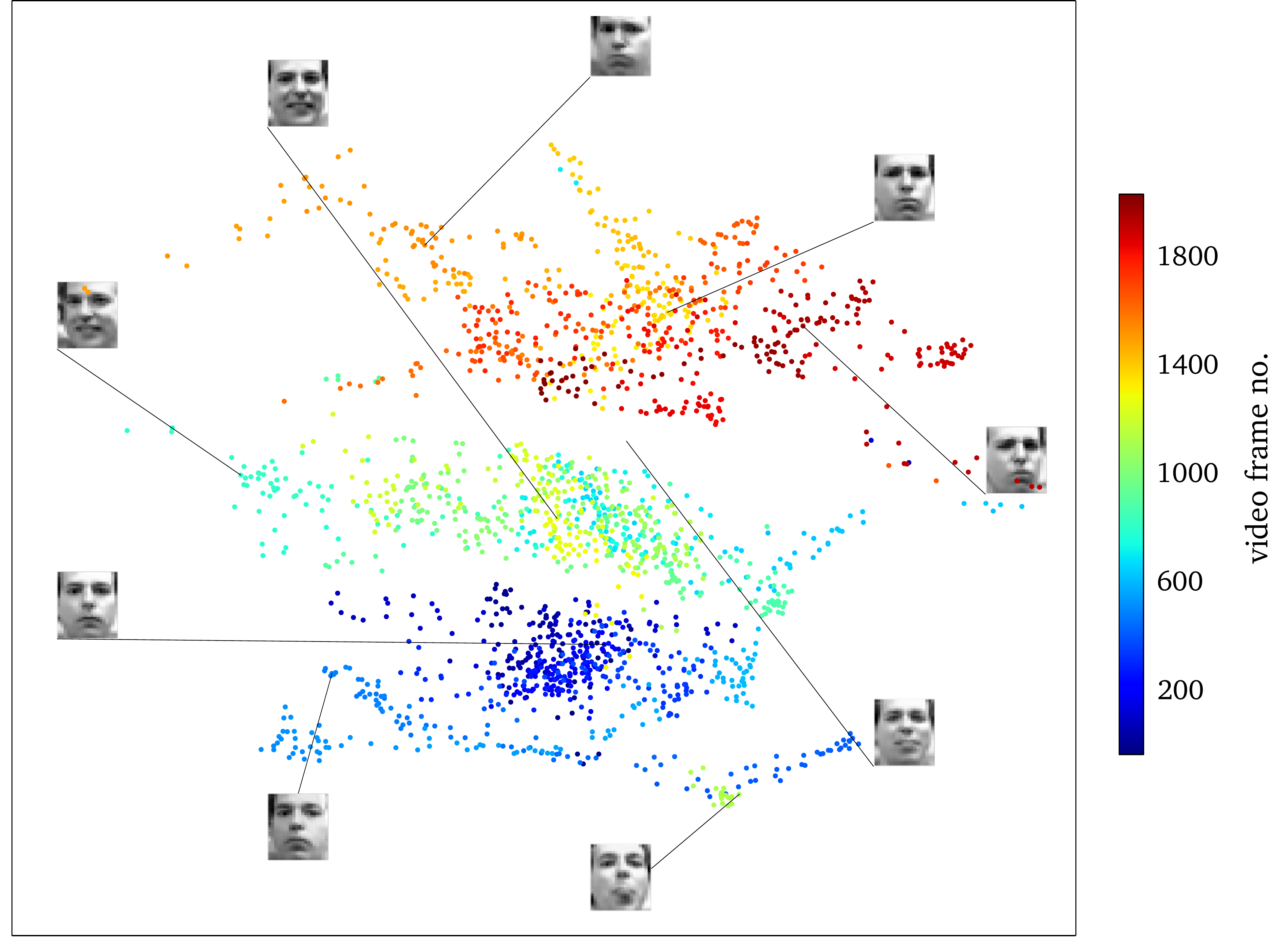}
  \caption{\small \label{fig:brendan} Latent space of the Frey faces data set
    using an autoencoding deep GP. We show illustrative latent points
    shown reconstructed. Each point is colored according to its
    position in the video, although the model is trained ignoring the
    temporal structure of the data.}
\end{figure}

\section{Discussion}
\label{sec:discuss}

We have shown how variational compression bounds can be applied to
inference in deep \shortlong{GP}{Gaussian process} models. The resulting bounds are
scalable due to their decompositional nature over the data points. In
analysing the bound on the deep GP marginal likelihood, we have seen
that the approximation appears as a neural network structure, but with
Gaussian messages being fed forward. We have shown how natural penalty
terms arise which decrease the bound when the propagation of the
Gaussian messages performs poorly.

This work opens the possibility of application of Deep GPs to big datasets for the first time. 
%

\newpage
\bibliography{lawrence,other,zbooks}

\begin{thebibliography}{30}
\providecommand{\natexlab}[1]{#1}
\providecommand{\url}[1]{\texttt{#1}}
\expandafter\ifx\csname urlstyle\endcsname\relax
  \providecommand{\doi}[1]{doi: #1}\else
  \providecommand{\doi}{doi: \begingroup \urlstyle{rm}\Url}\fi

\bibitem[\'Alvarez et~al.(2010)\'Alvarez, Luengo, Titsias, and
  Lawrence]{Alvarez:efficient10}
M.~A. \'Alvarez, D.~Luengo, M.~K. Titsias, and N.~D. Lawrence.
\newblock Efficient multioutput {G}aussian processes through variational
  inducing kernels.
\newblock In  \citet{Teh:aistats10}, pages 25--32.

\bibitem[Calandra et~al.(2014)Calandra, Peters, Rasmussen, and
  Deisenroth]{Calandra:manifold}
R.~Calandra, J.~Peters, C.~E. Rasmussen, and M.~P. Deisenroth.
\newblock Manifold {G}aussian processes for regression.
\newblock Technical report, 2014.

\bibitem[Csat\'o and Opper(2002)]{Csato:sparse02}
L.~Csat\'o and M.~Opper.
\newblock Sparse on-line {G}aussian processes.
\newblock \emph{Neural Computation}, 14\penalty0 (3):\penalty0 641--668, 2002.

\bibitem[Damianou and Lawrence(2013)]{Damianou:deepgp13}
A.~Damianou and N.~D. Lawrence.
\newblock Deep {G}aussian processes.
\newblock In C.~Carvalho and P.~Ravikumar, editors, \emph{Proceedings of the
  Sixteenth International Workshop on Artificial Intelligence and Statistics},
  volume~31, AZ, USA, 2013. JMLR W\&CP 31.

\bibitem[Damianou et~al.(2011)Damianou, Titsias, and
  Lawrence]{Damianou:vgpds11}
A.~Damianou, M.~K. Titsias, and N.~D. Lawrence.
\newblock Variational {Gaussian} process dynamical systems.
\newblock In P.~Bartlett, F.~Peirrera, C.~Williams, and J.~Lafferty, editors,
  \emph{Advances in Neural Information Processing Systems}, volume~24,
  Cambridge, MA, 2011. MIT Press.

\bibitem[Denil et~al.(2013)Denil, Shakibi, Dinh, Ranzato, and
  Freitas]{Denil:predicting13}
M.~Denil, B.~Shakibi, L.~Dinh, M.~Ranzato, and N.~D. Freitas.
\newblock Advances in neural information processing systems.
\newblock In C.~J.~C. Burges, L.~Bottou, Z.~Ghahramani, M.~Welling, and K.~Q.
  Weinberger, editors, \emph{Advances in Neural Information Processing
  Systems}, volume~26, pages 2148--2156, Cambridge, MA, 2013.

\bibitem[Duvenaud et~al.(2014)Duvenaud, Rippel, Adams, and
  Ghahramani]{Duvenaud:pathologies14}
D.~Duvenaud, O.~Rippel, R.~Adams, and Z.~Ghahramani.
\newblock Avoiding pathologies in very deep networks.
\newblock In S.~Kaski and J.~Corander, editors, \emph{Proceedings of the
  Seventeenth International Workshop on Artificial Intelligence and
  Statistics}, volume~33, Iceland, 2014. JMLR W\&CP 33.

\bibitem[Ferris et~al.(2007)Ferris, Fox, and Lawrence]{Ferris:wifi07}
B.~D. Ferris, D.~Fox, and N.~D. Lawrence.
\newblock {WiFi-SLAM} using {G}aussian process latent variable models.
\newblock In M.~M. Veloso, editor, \emph{Proceedings of the 20th International
  Joint Conference on Artificial Intelligence (IJCAI 2007)}, pages 2480--2485,
  2007.

\bibitem[Frey et~al.(1998)Frey, Colmenarez, and Huang]{Frey:mixtures98}
B.~J. Frey, A.~Colmenarez, and T.~S. Huang.
\newblock Mixtures of local linear subspaces for face recognition.
\newblock In \emph{Computer Vision and Pattern Recognition, 1998. Proceedings.
  1998 IEEE Computer Society Conference on}, pages 32--37. IEEE, 1998.

\bibitem[Gal et~al.(2014)Gal, van~der Wilk, and Rasmussen]{Gal:distributed14}
Y.~Gal, M.~van~der Wilk, and C.~E. Rasmussen.
\newblock Distributed variational inference in sparse {G}aussian process
  regression and latent variable models.
\newblock In Z.~Ghahramani, M.~Welling, C.~Cortes, N.~D. Lawrence, and K.~Q.
  Weinberger, editors, \emph{Advances in Neural Information Processing
  Systems}, volume~27, Cambridge, MA, 2014.

\bibitem[Hensman et~al.(2013)Hensman, Fusi, and Lawrence]{Hensman:bigdata13}
J.~Hensman, N.~Fusi, and N.~D. Lawrence.
\newblock {G}aussian processes for big data.
\newblock In A.~Nicholson and P.~Smyth, editors, \emph{Uncertainty in
  Artificial Intelligence}, volume~29. AUAI Press, 2013.

\bibitem[Hoffman et~al.(2012)Hoffman, Blei, Wang, and
  Paisley]{Hoffman:stochastic12}
M.~Hoffman, D.~M. Blei, C.~Wang, and J.~Paisley.
\newblock Stochastic variational inference.
\newblock Technical report, 2012.

\bibitem[Hornik et~al.(1989)Hornik, Stinchcombe, and White]{Hornik:univ89}
K.~Hornik, M.~Stinchcombe, and H.~White.
\newblock Multilayer feedforward networks are universal approximators.
\newblock \emph{Neural Networks}, 2\penalty0 (5):\penalty0 359--366, 1989.

\bibitem[Kingma and Welling(2013)]{Kingma:auto13}
D.~P. Kingma and M.~Welling.
\newblock Auto-encoding variational {B}ayes.
\newblock Technical report, 2013.

\bibitem[Lawrence(2000)]{Lawrence:thesis00}
N.~D. Lawrence.
\newblock \emph{Variational Inference in Probabilistic Models}.
\newblock PhD thesis, Computer Laboratory, University of Cambridge, New Museums
  Site, Pembroke Street, Cambridge, CB2 3QG, U.K., 2000.
\newblock Available from \url{http://www.thelawrences.net/neil}.

\bibitem[Lawrence and Moore(2007)]{Lawrence:hgplvm07}
N.~D. Lawrence and A.~J. Moore.
\newblock Hierarchical {G}aussian process latent variable models.
\newblock In Z.~Ghahramani, editor, \emph{Proceedings of the International
  Conference in Machine Learning}, volume~24, pages 481--488. Omnipress, 2007.
\newblock ISBN 1-59593-793-3.

\bibitem[L{\'a}zaro-Gredilla(2012)]{Lazaro:warped12}
M.~L{\'a}zaro-Gredilla.
\newblock {B}ayesian warped {G}aussian processes.
\newblock In P.~L. Bartlett, F.~C.~N. Pereira, C.~J.~C. Burges, L.~Bottou, and
  K.~Q. Weinberger, editors, \emph{Advances in Neural Information Processing
  Systems}, volume~25, Cambridge, MA, 2012.

\bibitem[{MacKay}(1998)]{MacKay:gpintroduction98}
D.~J.~C. {MacKay}.
\newblock Introduction to {G}aussian {P}rocesses.
\newblock In C.~M. Bishop, editor, \emph{Neural Networks and Machine Learning},
  volume 168 of \emph{Series F: Computer and Systems Sciences}, pages 133--166.
  Springer-Verlag, Berlin, 1998.

\bibitem[Neal(1996)]{Neal:book96}
R.~M. Neal.
\newblock \emph{{B}ayesian Learning for Neural Networks}.
\newblock Springer, 1996.
\newblock Lecture Notes in Statistics 118.

\bibitem[{Qui\~nonero Candela} and Rasmussen(2005)]{Quinonero:unifying05}
J.~{Qui\~nonero Candela} and C.~E. Rasmussen.
\newblock A unifying view of sparse approximate {G}aussian process regression.
\newblock \emph{Journal of Machine Learning Research}, 6:\penalty0 1939--1959,
  2005.

\bibitem[Rasmussen and Williams(2006)]{Rasmussen:book06}
C.~E. Rasmussen and C.~K.~I. Williams.
\newblock \emph{Gaussian Processes for Machine Learning}.
\newblock MIT Press, Cambridge, MA, 2006.
\newblock ISBN 0-262-18253-X.

\bibitem[Rezende et~al.(2014)Rezende, Mohamed, and
  Wierstra]{Rezende:stochastic14}
D.~J. Rezende, S.~Mohamed, and D.~Wierstra.
\newblock Stochastic back-propagation and variational inference in deep latent
  {G}aussian models.
\newblock Technical report, 2014.

\bibitem[Snelson and Ghahramani(2006)]{Snelson:pseudo05}
E.~Snelson and Z.~Ghahramani.
\newblock Sparse {G}aussian processes using pseudo-inputs.
\newblock In Y.~Weiss, B.~Sch\"olkopf, and J.~C. Platt, editors, \emph{Advances
  in Neural Information Processing Systems}, volume~18, Cambridge, MA, 2006.
  MIT Press.

\bibitem[Srivastava et~al.(2014)Srivastava, Hinton, Krizhevsky, Sutskever, and
  Salakhutdinov]{Srivastava:dropout14}
N.~Srivastava, G.~Hinton, A.~Krizhevsky, I.~Sutskever, and R.~Salakhutdinov.
\newblock Dropout: A simple way to prevent neural networks from overfitting.
\newblock \emph{Journal of Machine Learning Research}, 15:\penalty0 1929--1958,
  2014.
\newblock URL \url{http://jmlr.org/papers/v15/srivastava14a.html}.

\bibitem[Teh and Titterington(2010)]{Teh:aistats10}
Y.~W. Teh and D.~M. Titterington, editors.
\newblock \emph{Artificial Intelligence and Statistics}, volume~9, Chia Laguna
  Resort, Sardinia, Italy, 13-16 May 2010. JMLR W\&CP 9.

\bibitem[Titsias(2009)]{Titsias:variational09}
M.~K. Titsias.
\newblock Variational learning of inducing variables in sparse {G}aussian
  processes.
\newblock In D.~{van Dyk} and M.~Welling, editors, \emph{Proceedings of the
  Twelfth International Workshop on Artificial Intelligence and Statistics},
  volume~5, pages 567--574, Clearwater Beach, FL, 16-18 April 2009. JMLR W\&CP
  5.

\bibitem[Titsias and Lawrence(2010)]{Titsias:bayesGPLVM10}
M.~K. Titsias and N.~D. Lawrence.
\newblock Bayesian {G}aussian process latent variable model.
\newblock In  \citet{Teh:aistats10}, pages 844--851.

\bibitem[Williams(1998)]{Williams:computation98}
C.~K.~I. Williams.
\newblock Computation with infinite neural networks.
\newblock \emph{Neural Computation}, 10\penalty0 (5):\penalty0 1203--1216,
  1998.

\bibitem[Wilson and Adams(2013)]{Wilson:gpatt13}
A.~G. Wilson and R.~P. Adams.
\newblock Gaussian process kernels for pattern discovery and extrapolation.
\newblock In S.~Dasgupta and D.~McAllester, editors, \emph{ICML}, volume~28 of
  \emph{JMLR Proceedings}, pages 1067--1075. JMLR.org, 2013.

\bibitem[Zhu et~al.(1997)Zhu, Byrd, and Nocedal]{Zhu:lbfgsb97}
C.~Zhu, R.~H. Byrd, and J.~Nocedal.
\newblock {L-BFGS-B}: Algorithm 778: {L-BFGS-B}, {FORTRAN} routines for large
  scale bound constrained optimization.
\newblock \emph{ACM Transactions on Mathematical Software}, 23\penalty0
  (4):\penalty0 550--560, 1997.

\end{thebibliography}
\bibliographystyle{abbrvnat}

\end{document}